\documentclass[sigconf]{acmart}

\usepackage{makecell}
\usepackage{multirow}
\AtBeginDocument{%
  }

\copyrightyear{2026}
\acmYear{2026}
\setcopyright{cc}
\setcctype{by}
\acmConference[MM '26]{Proceedings of the 34th ACM International Conference on Multimedia}{November 10--14, 2026}{Rio de Janeiro, Brazil}
\acmBooktitle{Proceedings of the 34th ACM International Conference on Multimedia (MM '26), November 10--14, 2026, Rio de Janeiro, Brazil}
\acmDOI{10.1145/3767308.3835893}
\acmISBN{979-8-4007-2213-4/2026/11}

\begin{document}

\title{CityPaint: A Free Lunch from 2D Diffusion for City-Scale 3D Texturing}

\author{Yuchen Zou}
\authornote{Both authors contributed equally to this research.\protect\\
This work was completed during internship at LASER IDEA.}
\affiliation{%
  \institution{Xi'an Jiaotong University}
  \city{Xi'an}
  \country{China}}
\email{yuchenzou@stu.xjtu.edu.cn}

\author{Lihuang Fang}
\authornotemark[1]
\affiliation{%
  \institution{Southern University of Science and Technology}
  \city{Shenzhen}
  \country{China}}
\email{l.h.fang228@gmail.com}

\author{Siyu Zhu}
\affiliation{%
  \institution{Xi'an Jiaotong University}
  \city{Xi'an}
  \country{China}}
\email{zhusiyu@stu.xjtu.edu.cn}

\author{Xiao Hu}
\authornote{Corresponding author.}
\affiliation{%
  \institution{International Digital Economy Academy}
  \city{Shenzhen}
  \country{China}}
\affiliation{%
  \institution{HKUST (Guangzhou)}
  \city{Guangzhou}
  \country{China}}
\email{huxiao1@idea.edu.cn}

\renewcommand{\shortauthors}{Zou et al.}

\begin{abstract}
Generating high-fidelity, city-scale 3D environments plays a crucial role in applications such as virtual reality, gaming, and urban digital twins. However, the scarcity of real-world city-scale 3D training data forces existing methods to either stitch independently generated assets or learn from synthetic urban corpora, often sacrificing realism and global consistency. In contrast, 2D image data and their corresponding generative models are abundant and encode rich visual priors. To bridge this gap, we propose CityPaint, a novel framework that leverages 2D diffusion models to generate globally consistent city-scale 3D textures without relying on city-scale 3D training data. Specifically, CityPaint integrates geometric priors with the generative power of 2D diffusion models, introducing a robust top-$K$ camera baking strategy for iterative texture projection. To address the artifacts of 2D-to-3D projection, we further propose a decoupled refinement mechanism that separately optimizes terrain and building textures. Extensive experiments show that CityPaint produces globally coherent textures with fine local details and achieves favorable results against strong baselines in both visual quality and cross-view consistency.
\end{abstract}


\begin{teaserfigure}
  \centering
  \includegraphics[width=0.9\textwidth]{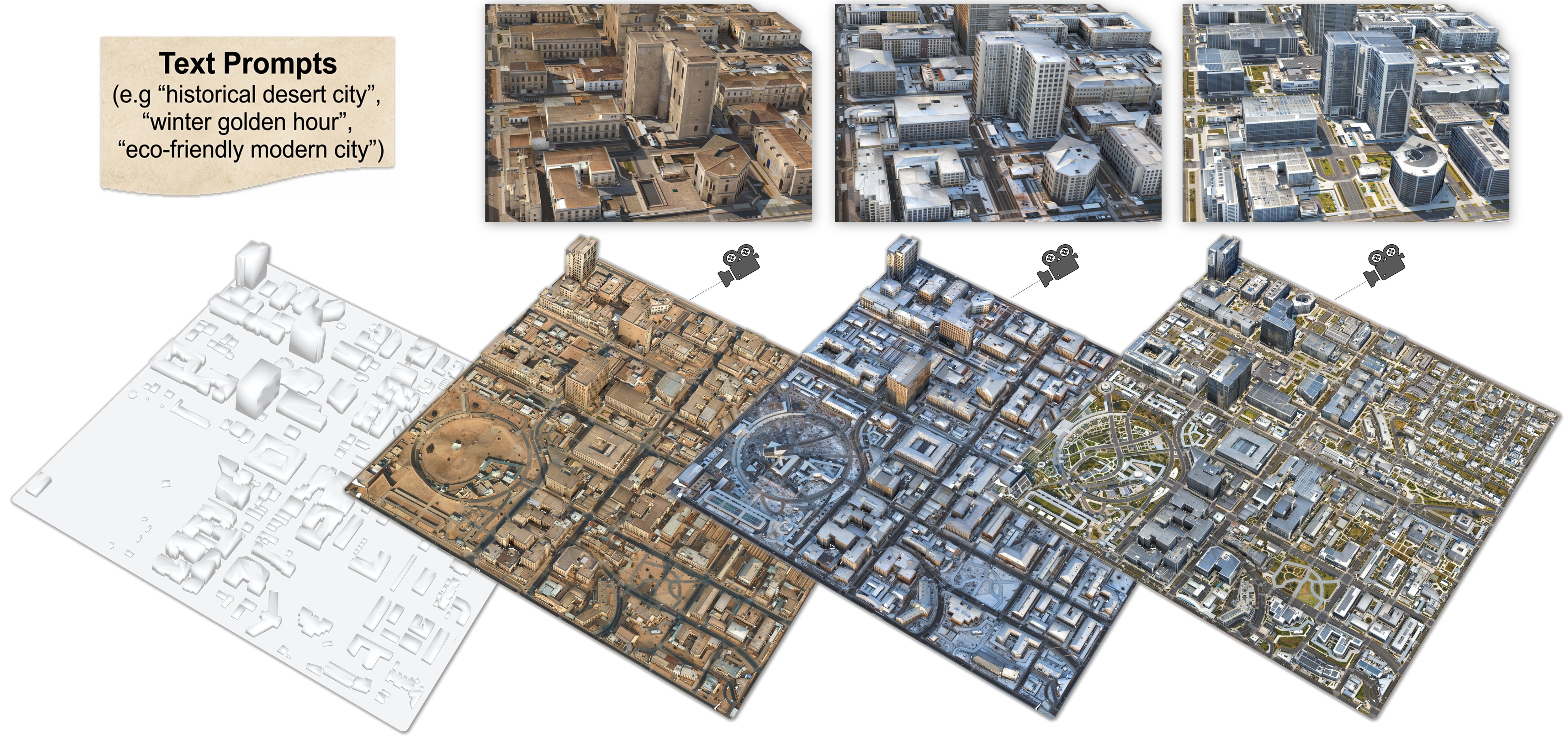}
  \caption{\textbf{Capabilities of CityPaint.} Given text prompts, our framework generates high-fidelity, globally consistent 3D city textures. By leveraging the rich visual priors of 2D diffusion models, CityPaint naturally supports diverse architectural styles, climates, and lighting conditions without relying on city-scale 3D training data.}
  \Description{Rendered city scenes with varied styles, climates, lighting, and viewpoints.}
  \label{fig:teaser}
\end{teaserfigure}


\maketitle

\section{Introduction}
\label{sec:intro}

The demand for expansive, high-fidelity 3D urban environments has surged in recent years, driven by rapidly expanding applications in virtual reality, immersive gaming, autonomous driving simulation, and urban digital twins \cite{proceduralcity2018,roboticcity2015,vrsmartcities2017,cityplanningvr2016}. Traditionally, constructing these large-scale scenes relies on exhaustive manual modeling by technical artists~\cite{scenedreamer2023}. While recent advancements in 3D generative AI have shown remarkable success in synthesizing single objects or constrained indoor scenes \cite{dreamfusion2022,magic3d2023,zero1232023,mvdream2023}, subsequent feed-forward and texturing-oriented systems have further improved consistency and efficiency \cite{wonder3d2023,roompainter2025,mvpaint2025}. Scaling these capabilities to the magnitude of an entire city nevertheless remains a formidable challenge.

The primary bottleneck in city-scale 3D generation lies in the extreme scarcity of high-quality, real-world urban 3D data. Constrained by this limitation, existing generative approaches are forced into a fundamental trade-off, typically falling into one of two suboptimal paradigms. On one hand, bottom-up instance composition methods synthesize individual buildings and subsequently stitch them into a scene \cite{citygen2023,citycraft2024,scenex2024,citysculpt2025,majutsucity2025}. While scalable, such approaches inevitably break global contextual consistency and compromise stylistic coherence. On the other hand, holistic scene-level models rely on synthetic datasets or low-quality satellite imagery for training \cite{infinicity2023,citydreamer2024,sat2scene2024,gaussiancity2025,magiccity2025}, which introduce significant domain gaps and inherit physical artifacts, such as distorted (``melted'') geometries and baked-in shadows. As a result, the generated environments often lack the richness and realism of real-world cities.

In stark contrast, while 3D urban data remains scarce, 2D image data is abundantly available. Modern diffusion models \cite{ldm2022,controlnet2023,esser2024scaling,crowson2024scalable,wu2025qwen,cai2025z} have demonstrated an extraordinary ability to capture globally consistent visual priors, encoding rich semantic structures of urban layouts and architectural styles. This raises a fundamental question: can we leverage these pre-trained 2D priors as a ``free lunch'' to enable photorealistic 3D city texturing without requiring city-scale 3D supervision?

To bridge this gap, we propose CityPaint, a 3D-data-free framework that reformulates city-scale 3D generation as a geometry-grounded generative reconstruction problem. Instead of synthesizing unconstrained 3D volumes from scratch, CityPaint leverages readily available geographic data, such as OpenStreetMap, to initialize a semantically structured city mesh. As illustrated in Fig.~\ref{fig:teaser}, this lightweight geometric scaffold serves as a structural canvas, onto which the generative power of pre-trained 2D diffusion models is projected, enabling diverse textures and stylistic variations directly on 3D surfaces.

However, directly projecting 2D generations onto large-scale 3D geometry introduces significant challenges. Fixed-height camera trajectories either collide with tall structures or drift excessively above low-rise regions, while oblique viewpoints lead to severe texture distortion and cross-view inconsistencies. To address these issues, CityPaint introduces a heightfield-aware camera placement strategy that adapts to local urban geometry, ensuring stable and consistent observations. Furthermore, rather than naively averaging multi-view projections, we propose a robust face-wise top-$K$ camera baking scheme. By explicitly modeling visibility and discarding grazing-angle observations, this strategy enables seamless texture aggregation into a globally coherent UV atlas, effectively avoiding the blurring artifacts common in conventional blending approaches.

To account for the distinct visual characteristics of different urban elements, we introduce a decoupled refinement mechanism. Large-scale top-down renderings often lack sufficient resolution to capture fine-grained facade details, while stitched terrain patches tend to introduce noticeable seams. These observations suggest that a uniform treatment of the entire city is inherently suboptimal. To address this, CityPaint decomposes the refinement process, optimizing each structural component in its most suitable representation.

In summary, our main contributions are as follows:
\begin{itemize}
    \item We propose \textbf{CityPaint}, a novel 3D-data-free framework that leverages the rich visual priors of pre-trained 2D diffusion models to generate high-fidelity and globally consistent city-scale 3D textures.
    \item We introduce a heightfield-aware observation policy, coupled with a robust top-$K$ camera baking strategy, which effectively resolves visibility conflicts and mitigates texture distortion in large-scale 2D-to-3D projection.
    \item We design a decoupled refinement mechanism that optimizes terrain in UV space and building facades in instance space, significantly improving both global continuity and fine-grained structural details.
    \item Extensive experiments show that CityPaint achieves stronger overall performance than the evaluated baselines in visual fidelity, realism, and cross-view consistency, offering a scalable solution for high-quality urban 3D content generation.
\end{itemize}

\begin{table}[t]
  \centering
  \caption{Comparison of existing 3D city generation paradigms.}
  \Description{Comparison of six methods across four training and control properties.}
  \label{tab:method_comparison}
  \small
  \begin{tabular}{lcccc}
    \toprule
    \textbf{Method} & \makecell{\textbf{3D Training}\\\textbf{Free}} & \makecell{\textbf{3D Dataset}\\\textbf{Free}} & \makecell{\textbf{Asset}\\\textbf{Free}} & \makecell{\textbf{Text}\\\textbf{Control}} \\
    \midrule
    CityDreamer~\cite{citydreamer2024} & $\times$ & $\times$ & $\checkmark$ & $\times$ \\
    UrbanWorld~\cite{urbanworld2024}&  $\times$ & $\times$ & $\checkmark$ & $\checkmark$ \\
    GaussianCity~\cite{gaussiancity2025} & $\times$ & $\times$ & $\checkmark$ & $\times$ \\
    CitySculpt~\cite{citysculpt2025} & $\times$ & $\times$ & $\checkmark$ & $\times$ \\
    MajutsuCity~\cite{majutsucity2025} & $\times$ & $\times$ & $\times$ & $\checkmark$ \\
    \textbf{CityPaint (Ours)} & \textbf{$\checkmark$} & \textbf{$\checkmark$} & \textbf{$\checkmark$} & \textbf{$\checkmark$} \\
    \bottomrule
  \end{tabular}
\end{table}

\section{Related Work}
\label{sec:related_work}

\subsection{2D-to-3D Generation via Diffusion Priors}
The emergence of large-scale 2D diffusion models has significantly advanced 3D content generation. A dominant line of work leverages Score Distillation Sampling (SDS) \cite{dreamfusion2022,magic3d2023,dreamgaussian2023,lgm2024} to lift 2D diffusion priors into 3D representations, including Neural Radiance Fields (NeRF) \cite{nerf2020}, explicit meshes, and 3D Gaussian Splatting (3DGS) \cite{gaussiansplatting2023}. In parallel, feed-forward and multi-view diffusion models \cite{zero1232023,viewdiff2024,mvdream2023,cat3d2024} have been proposed to improve generation efficiency and enforce cross-view consistency. Despite these advances, existing approaches are primarily designed for object-centric scenarios and struggle to scale to unbounded and structurally complex environments. In particular, unconstrained optimization in SDS-based methods often suffers from over-smoothing, mode collapse, and multi-face artifacts, while feed-forward approaches remain limited by their reliance on implicit geometric priors. These limitations become more pronounced in urban scenes, where large spatial extents, intricate structural layouts, and ambiguous text descriptions jointly exacerbate geometric inconsistencies and texture misalignment. Rather than relying on unstable gradient-based distillation, CityPaint grounds the generative process on deterministic geometry derived from semantic maps. This design effectively reduces spatial ambiguity and enables precise, globally consistent texture synthesis at the city scale.

\subsection{Urban Scene Generation}
Recent deep learning approaches for urban scene generation generally follow two paradigms. The first is instance-based composition, which employs object-level asset pipelines \cite{scenex2024,citysculpt2025, majutsucity2025} to generate individual buildings that are subsequently populated onto layouts sourced from OpenStreetMap \cite{openstreetmap2017}, satellite or generative models \cite{citygen2023,citycraft2024}. However, this bottom-up assembly naturally neglects the global urban context, resulting in disjointed environments with conflicting architectural styles. 

The second paradigm attempts holistic scene generation by training directly on large-scale 3D urban datasets. These approaches typically rely on either synthetic city corpora \cite{li2023matrixcity, magiccity2025, hua2025sat2city} or real-world captures such as satellite and Google Earth imagery \cite{infinicity2023, citydreamer2024, sat2scene2024,gaussiancity2025}. Unfortunately, both data sources present significant drawbacks. Models trained on synthetic data often exhibit a severe domain gap, lacking real-world intricacies. Conversely, pipelines driven by aerial photogrammetry inevitably inherit the physical flaws of their source data, suffering from ``melted'' geometries, blurry textures, and baked-in shadows. Our CityPaint framework bypasses these 3D data bottlenecks. By leveraging lightweight geographic data purely for geometric scaffolding, our method extracts photorealistic textures directly from 2D priors without requiring city-scale 3D urban training data.

\subsection{Texture Synthesis for 3D Meshes}
Recent methods for texturing existing 3D geometries primarily fall into two paradigms: direct UV-space generation and multi-view projection. Direct UV generation approaches \cite{texturify2022,pointuv2023,uvfree2024,texturegen2024} use generative models to synthesize unwrapped texture atlases. While subsequent coarse-to-fine and city-oriented extensions \cite{paint3d2024,citysculpt2025} have advanced this paradigm, their performance remains limited by the scarcity of high-quality, paired 3D-UV training data. Consequently, scaling these approaches to city-scale meshes is fundamentally impractical.

To overcome these data limitations, multi-view projection pipelines \cite{texture2023,text2tex2023,syncdiffusion2023,texpainter2024} have gained prominence. These methods bypass explicit UV generation by projecting synthesized 2D views directly onto the 3D mesh. Although synchronized and scene-level extensions \cite{syncmvd2024,texturedreamer2024,roompainter2025,mvpaint2025} have improved multi-view consistency, scaling these techniques to expansive urban environments introduces critical geometric challenges. Specifically, standard fixed-trajectory sampling struggles with extreme topography, while grazing-angle projections and naive view blending cause severe stretching, ghosting, and blurring. Alternatively, decomposing building facades into UVs \cite{urbanworld2024} introduces noticeable stitching artifacts and compromises realism. To overcome these challenges, we introduce a heightfield-aware top-$K$ camera baking strategy and a decoupled refinement mechanism tailored for complex urban geometries.

\section{Method}
\label{sec:method}

\begin{figure*}[t]
    \centering
    \includegraphics[width=\textwidth]{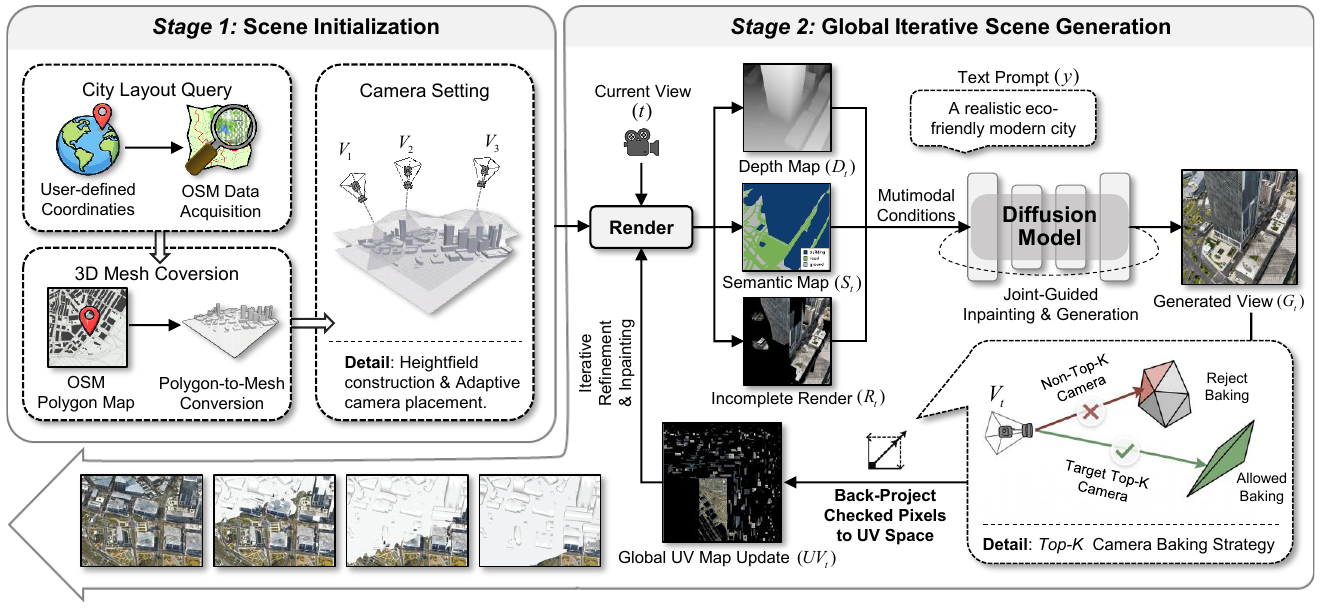}
    \caption{\textbf{Overview of the CityPaint Pipeline (Stages 1 \& 2).} In Stage 1, we construct a 3D city mesh from OpenStreetMap (OSM) data and establish heightfield-aware camera trajectories for stable observation. In Stage 2, a conditional diffusion model iteratively synthesizes the scene guided by rendered depth and semantic maps. The generated views are accumulated into a global UV atlas using a face-wise top-$K$ camera baking strategy to suppress artifacts from grazing angles.}
    \Description{Pipeline showing city mesh construction, adaptive camera sampling, conditional generation, and top-K texture baking.}
    \label{fig:citypaint_pipeline}
\end{figure*}

CityPaint generates city-scale textured 3D scenes from geographic coordinates and a text prompt. Instead of synthesizing urban environments from scratch, we exploit appearance priors from pre-trained 2D diffusion models and cast city texturing as a geometry-grounded generative reconstruction problem. As shown in Fig.~\ref{fig:citypaint_pipeline} and Fig.~\ref{fig:citypaint_refine}, the pipeline contains four stages: scene initialization, global iterative generation, terrain refinement, and building refinement.

\subsection{Scene Initialization}
\label{subsec:scene_init}

The first stage constructs the scene geometry and the observation policy used in later generation, as shown in the left part of Fig.~\ref{fig:citypaint_pipeline}. Given a user-specified city center $(\ell_{\mathrm{lat}}, \ell_{\mathrm{lng}})$ and retrieval radius $r$, we build a semantic city mesh from OSM building footprints and road layouts:
\begin{equation}
\mathcal{M} = \{\mathcal{M}^{\mathrm{bld}}, \mathcal{M}^{\mathrm{road}}, \mathcal{M}^{\mathrm{ter}}\},
\end{equation}
where $\mathcal{M}^{\mathrm{bld}}$, $\mathcal{M}^{\mathrm{road}}$, and $\mathcal{M}^{\mathrm{ter}}$ denote building, road, and terrain geometry, respectively. Specifically, we convert these map elements into a triangulated 3D scene by extruding building polygons and triangulating roads and terrain, then parameterize the textured region into a global UV atlas $\mathcal{U} \in [0,1]^{H_u \times W_u \times 3}$ that is updated throughout generation.

\paragraph{Heightfield construction.}
Given the initialized mesh $\mathcal{M}$, we project it onto the ground plane and estimate a heightfield over the horizontal domain $\Omega \subset \mathbb{R}^2$. For location $(x,z)$, the raw heightfield is defined as the upper envelope of triangles intersecting that ground position:
\begin{equation}
\label{eq:heightfield_raw}
H(x,z) = \max_{p \in \mathcal{M} \, : \, \pi(p) = (x,z)} p_y,
\end{equation}
where $\pi(p)$ projects a 3D point $p=(p_x,p_y,p_z)$ onto the $xz$-plane. In practice, we rasterize the scene into a grid and keep the maximum height per cell. This upper-envelope definition captures the dominant visible structure at each location. To suppress local spikes and smooth camera motion, we apply spatial smoothing:
\begin{equation}
\label{eq:heightfield_smooth}
\widetilde{H}(x,z) = (K * H)(x,z),
\end{equation}
where $K$ is an averaging kernel. The final operational heightfield is then lifted by a safety margin $\delta_h > 0$:
\begin{equation}
\label{eq:heightfield_lift}
H^{+}(x,z) = \widetilde{H}(x,z) + \delta_h.
\end{equation}

\paragraph{Adaptive camera placement.}
We sample scan trajectories that sweep across the scene and adapt the vertical position of each camera to the local geometry. Let $\gamma_t = (x_t, z_t)$ denote the 2D scanning location of the $t$-th camera. The corresponding 3D camera center is
\begin{equation}
\label{eq:camera_center}
\mathbf{c}_t = \left[x_t,\; H^{+}(x_t,z_t),\; z_t\right]^\top.
\end{equation}
The viewing direction is tilted downward by a fixed elevation angle $\theta$, producing a city-level oblique observation. 

\subsection{Global Iterative Scene Generation}
\label{subsec:global_generation}

The second stage textures the city through an iterative multiview generation loop. For each camera view $V_t$, we render a depth map $D_t$ and a semantic control map $S_t$:
\begin{equation}
\label{eq:render_controls}
(D_t, S_t) = \mathcal{R}_{\mathrm{cond}}(\mathcal{M}, V_t).
\end{equation}
Here $D_t$ provides geometry-aware distance cues, while $S_t$ encodes coarse semantic categories such as building, road, and terrain. Together they anchor diffusion sampling to both layout and region type.

At iteration $t$, the current UV atlas $\mathcal{U}_{t-1}$ is first rendered under the selected camera $V_t$ to obtain an incomplete observation
\begin{equation}
\label{eq:incomplete_render}
R_t = \mathcal{R}_{\mathrm{uv}}(\mathcal{U}_{t-1}, \mathcal{M}, V_t),
\end{equation}
where missing or insufficiently observed regions remain empty or unreliable. We then feed the text prompt $y$, rendered depth map $D_t$, semantic map $S_t$, and incomplete image $R_t$ into a conditional diffusion model:
\begin{equation}
\label{eq:global_generation}
G_t = \mathcal{G}_{\mathrm{global}}(y, D_t, S_t, R_t).
\end{equation}
The synthesized image $G_t$ is subsequently back-projected to UV space and fused into the global atlas. This loop turns view synthesis into an atlas completion process: each newly synthesized observation does not stand alone, but directly improves a persistent texture representation shared across all future views.

\paragraph{Iterative UV accumulation.}
For a visible pixel $q$ in view $V_t$, let $\phi_t(q) = (u_q, v_q)$ be its interpolated UV coordinate and let $\mathbf{g}_t(q) \in \mathbb{R}^3$ be the generated RGB value at that pixel. We distribute $\mathbf{g}_t(q)$ to its four neighboring UV texels using bilinear weights $\omega(q,p)$, $p \in \mathcal{N}(q)$. Denoting by $\mathcal{T}$ the set of UV texels, the running color sum $\mathbf{A}_t(p)$ and weight sum $W_t(p)$ for texel $p \in \mathcal{T}$ are updated as
\begin{equation}
\label{eq:uv_accum_sum}
\mathbf{A}_t(p) = \mathbf{A}_{t-1}(p) + \sum_{q \in \Pi_t(p)} \omega(q,p)\,\mathbf{g}_t(q),
\end{equation}
\begin{equation}
\label{eq:uv_accum_weight}
W_t(p) = W_{t-1}(p) + \sum_{q \in \Pi_t(p)} \omega(q,p),
\end{equation}
where $\Pi_t(p)$ collects view pixels whose bilinear neighborhood contains texel $p$. The atlas color is then obtained by normalized accumulation:
\begin{equation}
\label{eq:uv_normalize}
\mathcal{U}_t(p) = \frac{\mathbf{A}_t(p)}{\max(W_t(p), \epsilon)}.
\end{equation}

\paragraph{Face-wise top-$K$ camera baking.}
Directly baking every visible camera into the UV atlas often introduces severe artifacts, because highly oblique views produce stretched textures and conflicting observations. A central contribution of the global stage is therefore a face-wise top-$K$ camera baking strategy, as depicted in the bottom-right of Fig.~\ref{fig:citypaint_pipeline}. For each face $f$ and camera $V_t$, we define a visibility score based on the number of rendered pixels belonging to that face:
\begin{equation}
\label{eq:visibility_score}
s_{f,t}^{\mathrm{vis}} = \left| \{ q \in V_t \; | \; q \text{ is covered by } f \} \right|.
\end{equation}
Let $s_f^\star = \max_t s_{f,t}^{\mathrm{vis}}$ be the best visibility score for face $f$. We keep only the candidate views satisfying
\begin{equation}
\label{eq:topk_set}
\mathcal{K}_f =
\left\{
t \;\middle|\;
s_{f,t}^{\mathrm{vis}} \ge \rho \, s_f^\star,\;
t \in \mathrm{TopK}(f)
\right\},
\end{equation}
where $\rho \in (0,1]$ is a keep-ratio threshold and $\mathrm{TopK}(f)$ denotes the $K$ highest-scoring cameras for face $f$. During UV update, a pixel from camera $V_t$ is allowed to contribute to face $f$ only if $t \in \mathcal{K}_f$. This design suppresses low-quality grazing views while retaining a small set of complementary observations for hard-to-cover surfaces. In practice, it is this selective baking mechanism that makes globally coherent city-scale texturing feasible.

\begin{figure}[t]
    \centering
    \includegraphics[width=0.48\textwidth]{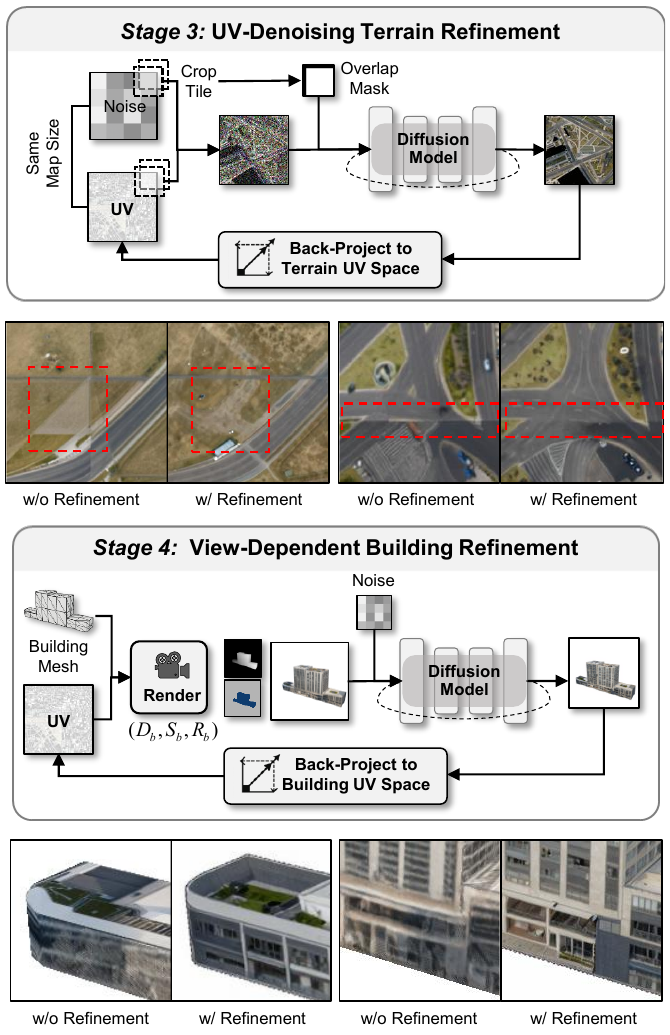}
    \caption{\textbf{Texture Refinement (Stages 3 \& 4).} Top: tile-wise UV denoising restores terrain continuity and removes seams. Bottom: object-centric orbit rendering recovers fine-grained building facade details.}
    \Description{Diagram showing terrain UV denoising and building facade refinement.}
    \label{fig:citypaint_refine}
\end{figure}

\subsection{UV-Denoising Terrain Refinement}
\label{subsec:terrain_refinement}
While global generation establishes the scene's macro-structure, top-$K$ baking can still leave fragmented patches over large terrain regions. Directly inpainting these regions often causes visible transitions. Since terrain elements such as roads and sidewalks occupy continuous UV regions and are less view-dependent than facades, we refine them directly in texture space to restore continuity (Fig.~\ref{fig:citypaint_refine}, top).

\paragraph{Terrain ROI extraction.}
Let $\Omega_{\mathrm{ter}} \subset \mathcal{T}$ denote the terrain-related UV region extracted from ground and road meshes. Rather than refining the entire atlas, we operate on the cropped terrain texture:
\begin{equation}
\label{eq:terrain_crop}
\mathcal{U}^{\mathrm{ter}} = \mathcal{U}_T \big|_{\Omega_{\mathrm{ter}}},
\end{equation}
where $T$ is the final iteration index of the global stage.

\paragraph{Tile-wise latent refinement.}
We divide $\Omega_{\mathrm{ter}}$ into overlapping tiles $\{\Omega_{\mathrm{ter}}^{(m)}\}_{m=1}^{M}$. A critical challenge in tile-wise generation is maintaining structural consistency across seams. Motivated by recent works demonstrating the spatial equivariance from noise space to image space \cite{chang2024warp, zhou2025alias, burgert2025go}, we pre-sample a global noise map $\mathbf{N} \sim \mathcal{N}(0,\mathbf{I})$ that shares the exact spatial dimensions as the full terrain UV latent map.

For the $m$-th tile, we simultaneously crop the current VAE latent $\mathbf{z}_0^{(m)}$ and its corresponding spatial noise patch $\mathbf{N}^{(m)}$. We then inject noise at a timestep determined by the refinement strength $\alpha \in [0,1]$:
\begin{equation}
\label{eq:terrain_noising}
\mathbf{z}_{\tau}^{(m)} =
\sqrt{\bar{\alpha}_{\tau}}\, \mathbf{z}_0^{(m)}
+
\sqrt{1-\bar{\alpha}_{\tau}}\, \mathbf{N}^{(m)},
\end{equation}
where $\tau$ is the strength-controlled starting timestep. By anchoring the noise to the global UV coordinates, we guarantee that overlapping regions share identical initial noise states. The denoising model then predicts a refined latent guided by a terrain-specific prompt $y_{\mathrm{ter}}$ and an overlap mask $M^{(m)}$ indicating the specific regions designated for regeneration:
\begin{equation}
\label{eq:terrain_refine}
\widehat{\mathbf{z}}^{(m)} =
\mathcal{G}_{\mathrm{ter}}\!\left(\mathbf{z}_{\tau}^{(m)}, y_{\mathrm{ter}}, M^{(m)}\right).
\end{equation}

 After decoding the refined latents $\widehat{\mathbf{z}}^{(m)}$ into image-space tiles $\widehat{\mathcal{U}}^{(m)}$, we directly update the full atlas by replacing the original texels within the masked regions:
\begin{equation}
\label{eq:terrain_merge}
\mathcal{U}_{T}^{+}(p) = \widehat{\mathcal{U}}^{(m)}(p), \quad \forall p \in \Omega_{\mathrm{ter}}^{(m)} \text{ s.t. } M^{(m)}(p) = 1.
\end{equation}

\subsection{View-Dependent Building Refinement}
\label{subsec:building_refinement}

The global generation stage relies on large-scale overhead observations, which inevitably introduce geometric misalignment and blurring on building facades due to long-range projections and steep viewing angles. To mitigate this, we process each building independently using localized orbital cameras that provide close-up geometric guidance (Fig.~\ref{fig:citypaint_refine}, Stage 4).

\paragraph{Object-centric orbit rendering.}
Let $\mathcal{B} = \{B_1, \dots, B_N\}$ denote the building instances extracted from the city mesh. For each building $B_n$, we sample a set of orbit cameras $\{V_{n,j}\}_{j=1}^{J_n}$ across various azimuths and elevations. From each view, we render a building-focused depth map $D_{n,j}$, a semantic map $S_{n,j}$, and the current global appearance $R_{n,j}$:
\begin{equation}
\label{eq:building_render}
(D_{n,j}, S_{n,j}, R_{n,j}) =
\mathcal{R}_{\mathrm{bld}}(B_n, \mathcal{U}_{T}^{+}, V_{n,j}).
\end{equation}
This localized setup outperforms global views by observing buildings from favorable angles and focusing generative capacity on local structures.

\paragraph{Building Refinement.}
Conditioned on the building prompt $y_{\mathrm{bld}}$ and the rendered control maps, the building generator synthesizes a refined facade observation:
\begin{equation}
\label{eq:building_generation}
G_{n,j} = \mathcal{G}_{\mathrm{bld}}(y_{\mathrm{bld}}, D_{n,j}, S_{n,j}, R_{n,j}).
\end{equation}
This ensures that local facade structures remain sharp and visually plausible. As demonstrated in the bottom row of Fig.~\ref{fig:citypaint_refine}, this stage effectively recovers high-frequency details (e.g., windows and architectural edges).

\paragraph{Normal-aware camera selection and update.}
For a building face $f$ observed under an orbit camera $V_{n,j}$, we define a geometric quality score using the cosine similarity between the face normal $\mathbf{n}_f$ and the viewing direction $\mathbf{v}_{f,j}$:
\begin{equation}
\label{eq:building_score}
s_{f,j}^{\mathrm{bld}} = \max\!\left(0, \mathbf{n}_f^\top \mathbf{v}_{f,j}\right).
\end{equation}
The valid view set for face $f$ is defined as:
\begin{equation}
\label{eq:building_topk}
\mathcal{K}_f^{\mathrm{bld}} =
\left\{
j \;\middle|\;
s_{f,j}^{\mathrm{bld}} \ge \eta \max_{j'} s_{f,j'}^{\mathrm{bld}},\;
j \in \mathrm{TopK}_{\mathrm{bld}}(f)
\right\},
\end{equation}
where $\eta$ is a threshold ratio. Only views within $\mathcal{K}_f^{\mathrm{bld}}$ are used to update the atlas. To preserve the sharp details from these high-confidence object-centric observations, we form the final texture by directly assigning synthesized texels to their corresponding UV regions:
\begin{equation}
\label{eq:building_overwrite}
\mathcal{U}^{\mathrm{final}}(p) =
\begin{cases}
\widehat{\mathcal{U}}_{n,j}(p), & p \in \Omega_{n,j}^{\mathrm{write}}, \\
\mathcal{U}_{T}^{+}(p), & \text{otherwise},
\end{cases}
\end{equation}
where $\Omega_{n,j}^{\mathrm{write}}$ denotes the valid texels covered by the selected views. The resulting atlas successfully preserves the large-scale consistency established in the global stage while adding sharp facade details.

Through these four stages, CityPaint transforms geographic priors into a coherent, city-scale textured mesh, balancing global consistency, terrain continuity, and building-level realism. 

\section{Experiments}

\subsection{Experimental Details}
\label{subsec:experimental_details}

Across all three stages (global generation, terrain refinement, and building refinement), we adopt the open-source generative model Z-Image~\cite{cai2025z} as the shared diffusion prior. 

In the global stage, we iteratively texture the scene onto an $8192 \times 8192$ UV atlas from 320 views. The cameras follow our heightfield-aware trajectory with 8 scan strips and 40 views per strip, using a $48^\circ$ downward elevation and a $40^\circ$ field of view. For each selected view, we render a depth map, a semantic control map, and the current partial appearance from the UV atlas, and run 25 denoising steps in the conditional diffusion model. 

For the subsequent refinement stages, terrain refinement operates on overlapping $1024 \times 1024$ UV tiles with a denoising strength of 0.2 and the shared global-noise initialization described in Eq.~\ref{eq:terrain_noising}. Building refinement uses object-centric orbit cameras sampled across azimuths at elevations of $20^\circ$ and $35^\circ$ to enhance facade details, and updates the atlas through the normal-aware selection rule in Eq.~\ref{eq:building_topk} with at most two views per face. All experiments are conducted on a single NVIDIA A100 GPU (40\,GB).

\begin{figure}[t]
  \centering
  \includegraphics[width=\linewidth]{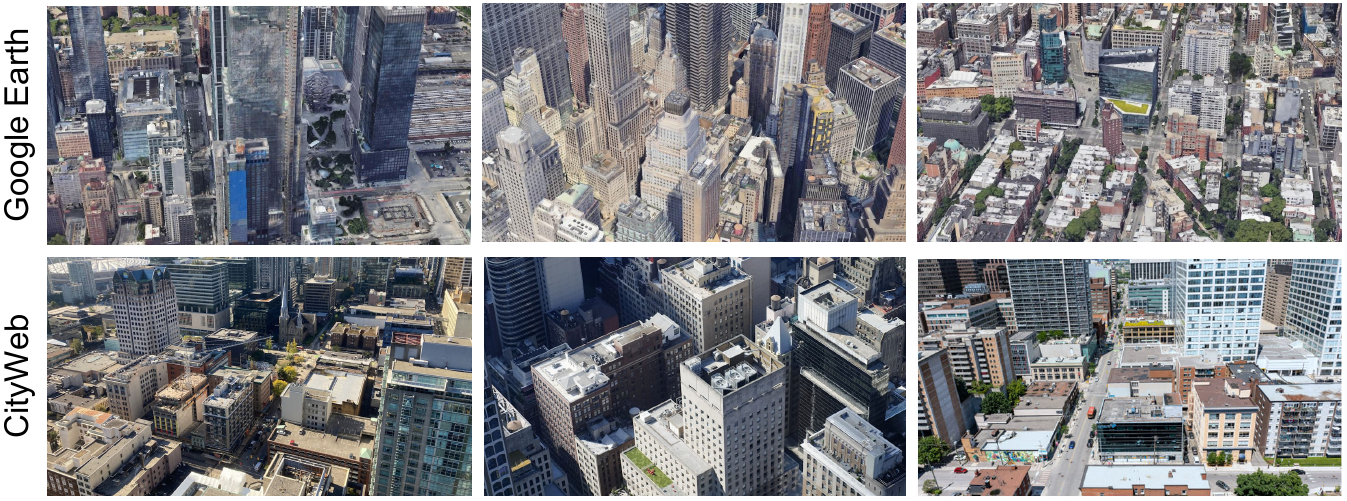} 
  \caption{Visual comparison of reference distributions. Top: Google Earth imagery often suffers from 3D reconstruction artifacts, blurred textures, and compression defects. Bottom: Our proposed \textbf{CityWeb} benchmark provides a cleaner, high-fidelity, and photorealistic reference distribution.}
  \Description{Blurred Google Earth examples above and sharper CityWeb photographs below.}
  \label{fig:cityweb_comparison}
\end{figure}

\begin{figure*}[t]
    \centering
    \includegraphics[width=0.96\textwidth]{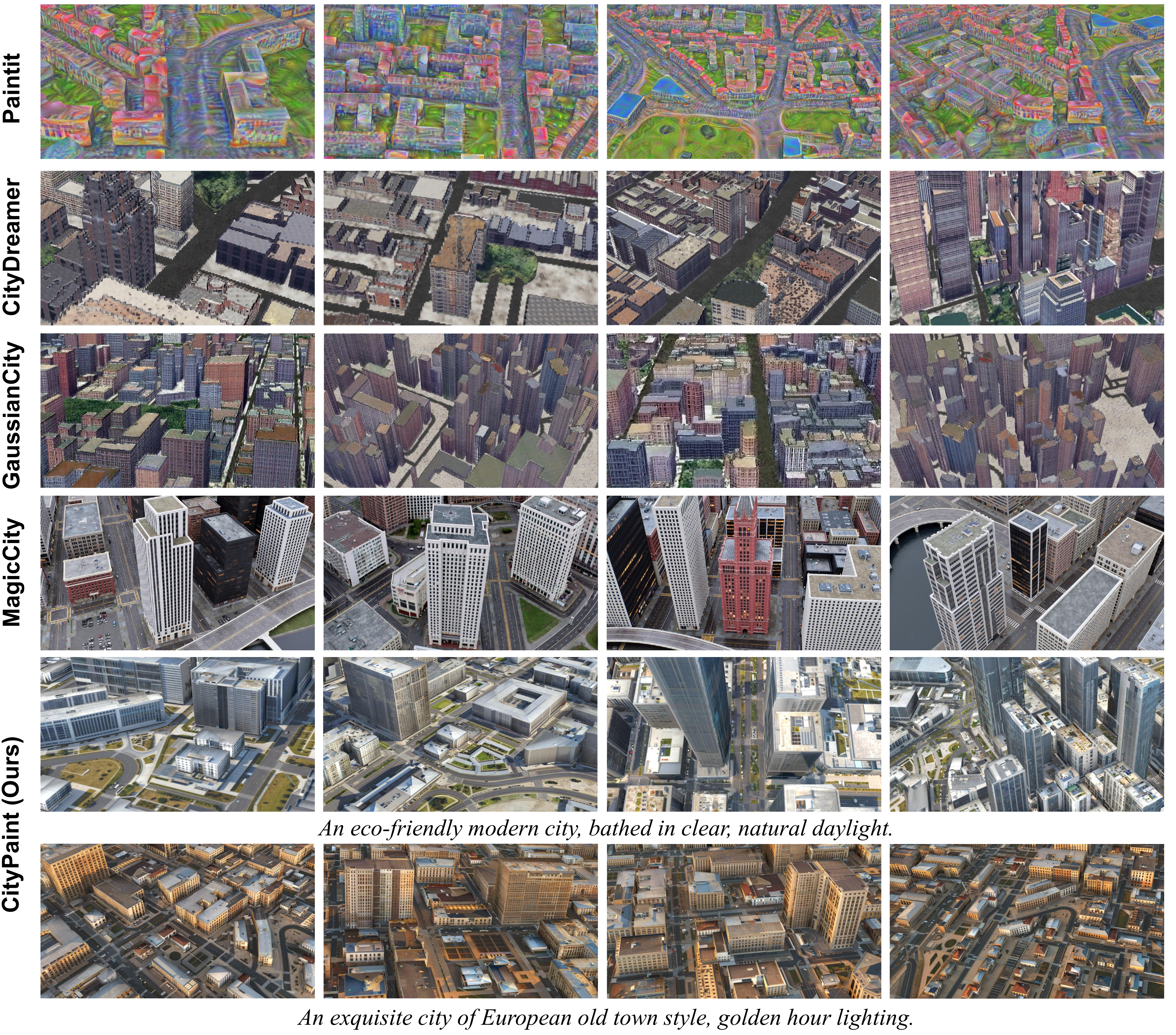}
    \caption{Qualitative comparison of different city generation methods (zoom-in recommended for details).}
    \Description{Grid comparing city scenes generated by baseline methods and CityPaint.}
    \label{fig:qualitative_comparison}
\end{figure*}

\begin{table*}
\centering
\caption{Quantitative comparison on geometric consistency (CE, DE) and visual fidelity (FID, KID, NIQE, BRISQUE) using Google Earth and CityWeb.}
\Description{Comparison of seven methods on geometric consistency and visual quality metrics.}
\label{tab:quantitative_comparison}
\begin{tabular}{l|c|c|c|cc|cc|c|c}
\toprule
\multirow{2}{*}{Method} & \multirow{2}{*}{Venue} & \multirow{2}{*}{CE $\downarrow$} & \multirow{2}{*}{DE $\downarrow$} & \multicolumn{2}{c|}{Google Earth} & \multicolumn{2}{c|}{CityWeb} & \multirow{2}{*}{NIQE $\downarrow$} & \multirow{2}{*}{BRISQUE $\downarrow$} \\
\cmidrule(lr){5-6}\cmidrule(lr){7-8}
& & & & FID $\downarrow$ & KID $\downarrow$ & FID $\downarrow$ & KID $\downarrow$ & & \\
\midrule
Text2Tex~\cite{text2tex2023} & ICCV'23 & 0.066 & 0.158 & 110.7 & 0.103 & 108.7 & 0.118 & 8.85 & 89.12 \\
Paintit~\cite{youwang2024paint} & CVPR'24 & 0.075 & 0.180 & 121.2 & 0.110 & 98.4 & 0.105 & 9.54 & 95.41 \\
CityDreamer~\cite{citydreamer2024} & CVPR'24 & 0.060 & 0.147 & 97.38 & 0.096 & 145.7 & 0.130 & 8.51 & 86.70 \\
GaussianCity~\cite{gaussiancity2025} & CVPR'25 & 0.057 & 0.136 & 86.94 & 0.090 & 125.4 & 0.137 & 7.69 & 75.34 \\
CitySculpt~\cite{citysculpt2025} & ACMMM'25 & 0.043 & \textbf{0.047} & 53.96 & 0.063 & - & - & 5.13 & 53.96 \\
MagicCity~\cite{magiccity2025} & ICCV'25 & 0.137 & 0.072 & - & - & - & - & 4.55 & 28.02 \\
CityPaint (Ours) & - & \textbf{0.035} & 0.051 & \textbf{49.21} & \textbf{0.055} & \textbf{42.84} & \textbf{0.044} & \textbf{3.28} & \textbf{17.15} \\
\bottomrule
\end{tabular}
\end{table*}

\subsection{Evaluation Metrics}
\label{subsec:evaluation_metrics}

A key challenge in evaluating city-scale 3D generation is the lack of reliable reference image distributions. Existing works commonly use Google Earth imagery~\cite{citydreamer2024, gaussiancity2025} as an evaluation benchmark. However, such imagery is not ideal for assessing photorealism because it often contains reconstruction artifacts, compression noise, and noticeable blurring, as illustrated in Fig.~\ref{fig:cityweb_comparison}. Consequently, metrics computed solely on this distribution may not accurately reflect the visual quality of generated results.

To address this limitation, we introduce a complementary high-quality benchmark, termed \textbf{CityWeb}. We collect a large-scale set of real-world urban photographs from publicly accessible web sources and apply a multi-stage filtering pipeline to remove non-photographic, low-resolution, and watermarked images. A GPT-based scoring mechanism~\cite{achiam2023gpt} then ranks the candidates according to urban relevance, visual clarity, and overall fidelity. The top 20,000 images are retained to form the CityWeb benchmark, yielding a cleaner and more realistic reference distribution than Google Earth alone.

We evaluate CityPaint using both benchmarks. For distribution-level alignment, we report Fr\'echet Inception Distance (FID) and Kernel Inception Distance (KID) against Google Earth for comparability with prior work and against CityWeb for high-fidelity realism. To further evaluate perceptual quality without relying on a reference distribution, we also report the no-reference metrics Natural Image Quality Evaluator (NIQE) and BRISQUE.

Finally, to assess the geometric consistency of the generated 3D scenes, we follow prior works~\cite{citydreamer2024, gaussiancity2025, citysculpt2025} and report two metrics. Camera Error (CE) measures the scale-invariant L2 distance between the inferred camera trajectory and poses estimated by COLMAP. Depth Error (DE) evaluates geometric accuracy using the normalized L2 distance between predicted and pseudo ground-truth depth maps.

\subsection{Evaluation and Comparisons}
\label{subsec:evaluation_comparisons}

We evaluate CityPaint against state-of-the-art baselines, with quantitative metrics summarized in Table~\ref{tab:quantitative_comparison} and qualitative visual results illustrated in Fig.~\ref{fig:qualitative_comparison}. To ensure a fair evaluation, we augment the SDS-based baseline Paintit~\cite{youwang2024paint} with semantic map guidance so that it can operate under the same urban layout constraints as our method. Because the full source code for CitySculpt~\cite{citysculpt2025} and MagicCity~\cite{magiccity2025} is not publicly available, we only report the metrics disclosed in their original publications and mark unavailable entries with ``-''.

\textbf{Image Quality \& Distribution Alignment:} CityPaint achieves the best overall results among the methods with complete evaluation in Table~\ref{tab:quantitative_comparison}. Although certain baselines report competitive FID scores on the Google Earth benchmark, that benchmark is itself affected by reconstruction artifacts and compression noise. Objective, reference-free metrics such as NIQE and BRISQUE provide a complementary view of the underlying blurring and synthetic artifacts, and the advantage of our method becomes clearer on the higher-fidelity CityWeb benchmark. These results indicate that leveraging strong 2D diffusion priors helps avoid inheriting artifact distributions from noisy urban data.

\textbf{Geometric \& Multi-view Consistency:} In addition to visual quality, our method demonstrates strong performance in preserving structural integrity and multi-view consistency. Direct 2D-to-3D projection at a city scale typically introduces severe texture stretching and viewpoint conflicts due to complex urban topographies. CityPaint effectively mitigates these geometric challenges through its heightfield-aware observation policy and robust top-$K$ camera baking strategy. Consequently, our approach yields view-consistent urban textures that remain well aligned with the complex 3D topology.

\textbf{Qualitative Comparisons:} As illustrated in Fig.~\ref{fig:qualitative_comparison}, CityPaint delivers stronger photorealism and prompt adherence than the compared baselines. Existing methods exhibit distinct failure modes: Paintit suffers from structural collapse, while CityDreamer and GaussianCity generate muddy, low-resolution facades. Additionally, MagicCity often yields a synthetic, plastic-like appearance. In contrast, CityPaint synthesizes more coherent cityscapes with sharper architectural details, such as window grids and reflective glass structures, while preserving the global style described by the prompt.

\begin{table}[t]
\centering
\caption{Ablation study of core components.}
\Description{Ablation results for the four core components of CityPaint.}
\label{tab:ablation_study}
\resizebox{\linewidth}{!}{
\begin{tabular}{l|cc|cc|cc}
\toprule
\multirow{2}{*}{Model Variations} & \multicolumn{2}{c|}{Geometry} & \multicolumn{2}{c|}{CityWeb} & \multicolumn{2}{c}{Perceptual} \\
\cmidrule(lr){2-3}\cmidrule(lr){4-5}\cmidrule(lr){6-7}
& CE $\downarrow$ & DE $\downarrow$ & FID $\downarrow$ & KID $\downarrow$ & NIQE $\downarrow$ & BRISQUE $\downarrow$ \\
\midrule
w/o Height-aware Cam. & 0.058 & 0.082 & 68.53 & 0.071 & 4.96 & 32.41 \\
w/o Top-$K$ Baking & 0.042 & 0.061 & 75.42 & 0.083 & 5.12 & 38.65 \\
w/o Terrain Refinement & 0.038 & 0.053 & 60.15 & 0.065 & 4.51 & 28.92 \\
w/o Building Refinement & 0.039 & 0.056 & 56.78 & 0.060 & 4.18 & 26.54 \\
\midrule
\textbf{CityPaint (Full)} & \textbf{0.035} & \textbf{0.051} & \textbf{42.84} & \textbf{0.044} & \textbf{3.28} & \textbf{17.15} \\
\bottomrule
\end{tabular}
}
\end{table}

\subsection{Ablation Study}
\label{subsec:ablation_study}

\begin{figure}[t]
\centering
\includegraphics[width=0.98\linewidth]{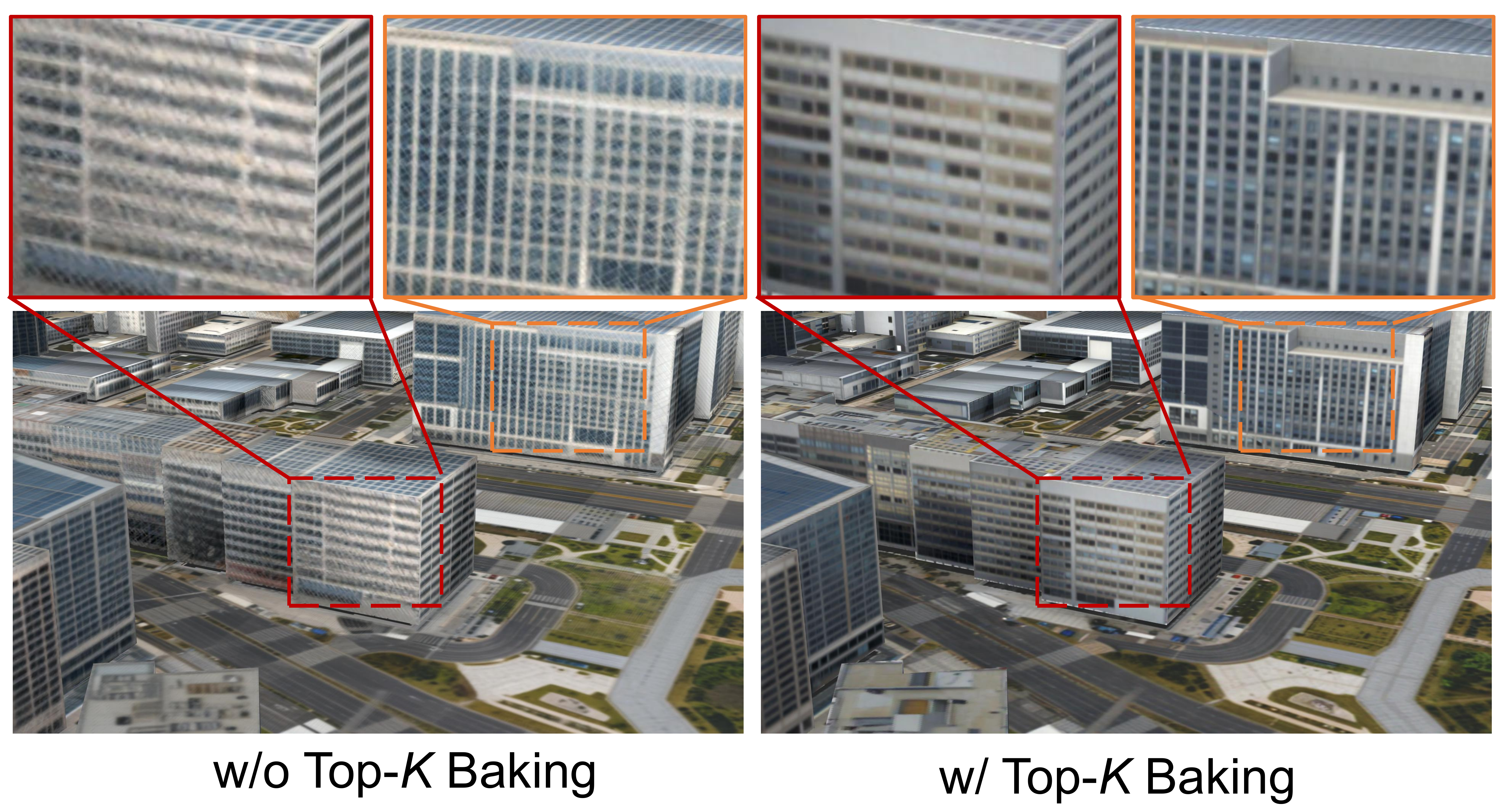}
\caption{Effect of top-$K$ camera baking. Without it (left), facade textures are blurred and inconsistent; our method (right) produces sharper, coherent details.}
\Description{Blurred facades without top-K baking on the left and sharper results with it on the right.}
\label{fig:ablation_topk}
\end{figure}

We ablate each core component to assess its impact on geometric consistency and image quality. Quantitative results are reported in Table~\ref{tab:ablation_study}.

\textbf{Effect of Heightfield-aware Camera Placement:} Replacing our adaptive trajectory with fixed-height grid observations (\textit{w/o Height-aware Cam.}) increases both CE and DE. Fixed cameras encounter grazing views or geometry collisions in uneven urban regions, causing texture stretching and projection mismatches.

\textbf{Effect of Top-$K$ Camera Baking:} Replacing face-wise view selection with multi-view averaging (\textit{w/o Top-$K$ Baking}) causes the largest drop in visual fidelity. Averaging conflicting observations introduces blurring, ghosting, and incoherent facades, whereas top-$K$ baking preserves sharp, consistent textures (Fig.~\ref{fig:ablation_topk}).

\textbf{Effect of Terrain Refinement:} Removing tile-wise UV denoising (\textit{w/o Terrain Refinement}) degrades perceptual quality because global projections leave large terrain regions at low effective resolution with visible tile-boundary seams. UV-space refinement restores coherent ground details.

\textbf{Effect of Building Refinement:} Removing instance-centric optimization (\textit{w/o Building Refinement}) increases FID and perceptual errors. Object-centric orbit views provide the pixel density needed to recover high-frequency facade details that are missing from global top-down observations.

\section{Conclusion}
\label{sec:conclusion}

We presented \textbf{CityPaint}, a 3D-data-free framework that reformulates city-scale 3D generation as a geometry-grounded generative reconstruction problem. By exploiting 2D diffusion priors, CityPaint reduces the reliance on scarce city-scale 3D urban datasets. Through face-wise top-$K$ camera baking and a decoupled refinement pipeline, our method synthesizes text-guided urban environments that balance global consistency with high-fidelity local details. Extensive evaluations show that CityPaint achieves stronger overall results than the compared baselines across geometric consistency and visual-quality metrics. Furthermore, we introduce \textbf{CityWeb}, a high-fidelity real-image benchmark designed to provide a more reliable reference for evaluating photorealism in urban scene synthesis.

\begin{acks}
This work was carried out during an internship at the Low Airspace Economy Research Institute (LASER), International Digital Economy Academy (IDEA). The support, resources, and research environment provided throughout the project are gratefully acknowledged.
\end{acks}

\bibliographystyle{ACM-Reference-Format}
\bibliography{sample-base}










\end{document}